\documentclass[11pt]{article}

\usepackage[final]{acl}

\usepackage{times}
\usepackage{latexsym}

\usepackage[T1]{fontenc}

\usepackage[utf8]{inputenc}

\usepackage{microtype}

\usepackage{inconsolata}

\usepackage{graphicx}
\usepackage{tabularx}
\usepackage{booktabs}
\usepackage{xurl}

\usepackage{xcolor}

\newcommand{\DEVELOPMENT}{0} 
\usepackage{ifthen}
\ifthenelse{\DEVELOPMENT = 1}{	
	\newcommand{\ah}[1]{\textcolor{cyan}{\textbf{AH:} #1}}	
    \newcommand{\js}[1]{\textcolor{blue}{\textbf{JS:} #1}}
   \newcommand{\todo}[1]{\textcolor{red}{\textbf{TODO:} #1}}
	\newcommand{\tbd}[1]{\textcolor{orange}{\textbf{TBD:} #1}}
	\newcommand{\outline}[1]{\textcolor{green}{\textbf{OUTLINE:} #1}}
}{
\newcommand{\ah}[1]{}	
\newcommand{\js}[1]{}
\newcommand{\todo}[1]{}
\newcommand{\tbd}[1]{}
\newcommand{\outline}[1]{}
}

\title{Detecting AI-Authored Text in Authentic Student Writing: Evidence from PISA 2025}

\author{Nils-Jonathan Schaller\textsuperscript{1}, \\
  \textbf{Thorben Jansen\textsuperscript{1}, Olaf Köller\textsuperscript{1}, Andrea Horbach\textsuperscript{1,2}} \\
  \textsuperscript{1}Leibniz Institute for Science and Mathematics Education, Kiel, Germany \\
  \textsuperscript{2}Kiel University, Kiel, Germany \\
  \texttt{schaller@leibniz-ipn.de} \\}

\begin{document}
\maketitle

\begin{abstract}
AI-text detectors reach near-perfect scores in evaluations that contrast fully human-written texts with texts generated from researcher-designed prompts. In authentic student writing, however, AI use is voluntary and often partial. We study this setting with the PISA 2025 FLA dataset, in which 15-year-olds in Germany wrote English argumentative essays with and without access to a \texttt{gpt-4o-mini}-based chatbot. Keystroke, clipboard and chat logs let us trace every character of the final essay to its origin. From this provenance, we derive text- and sentence-level labels for human-written, AI-generated and mixed text, and for text written after chatbot use without traceable AI material. We evaluate a feature-based detector and a classifier on frozen Llama-3.1-8B embeddings, both adapted from top-ranked shared-task systems. Both separate AI from human text reliably (macro-F1 up to 0.96 for texts and 0.95 for sentences), independently of spelling errors. The feature-based detector, however, relies on lexical complexity, which a simple student-style rewriting prompt removes (AI recall 0.92 $\to$ 0.60 with \texttt{gpt-4o-mini}, 0.41 with Qwen3). The embedding-based detector stays robust for whole texts (AI recall $\geq$ 0.94) but not for single sentences (AI recall as low as 0.59). Mixed text falls between the two classes, and text without traceable AI material is classified as human-written in about 95\% of texts. 
\end{abstract}

\section{Introduction}
Generative AI has become part of how students write: surveys report widespread use of chatbots for homework and writing assignments. In Germany, 57\% of 12- to 19-year-olds had used ChatGPT by 2024, and two thirds of those using AI tools do so for school and homework \citep{jim2024}; surveys in the US show a similar trend \citep{pew2025}.

For writing assessment, this raises a validity problem: if a text may have been produced partly by a chatbot, it no longer shows reliably what the student can do, and teachers find it hard to tell students' own writing from generated text \citep{cotton2024chatting, fleckenstein2024teachers}. Automated detectors are therefore increasingly used in education \citep{turnitin2023}, although their errors are costly: a false positive amounts to an accusation of misconduct, and widely used detectors misclassify texts by non-native writers as AI-generated at high rates \citep{liang2023gpt}.

In published evaluations, such detectors perform very well: in the 2025 \emph{AI vs.\ Human} shared task on academic essays, the best systems reached macro-F1 above 0.98 \citep{chowdhury-etal-2025-genai}. These evaluations, however, compare fully human-written texts with texts that researchers generated from carefully designed prompts. 
Students’ AI use in the classroom might look different, as they decide themselves whether to use a chatbot, prompt it loosely, combine its output with their own writing, or use it only for ideas. Partial use also means that AI material may be confined to individual sentences, so a useful detector should not only decide \emph{whether} a text contains AI material but also \emph{where}. It is unclear how detectors behave on such texts, how they treat partial AI use, and whether students could evade them with simple means.

We study these questions on the PISA 2025 FLA dataset, in which students in Germany aged 15 wrote English argumentative essays with and without access to a chatbot based on \texttt{gpt-4o-mini}. Keystrokes, clipboard operations and chat interactions were logged, so we can trace every character of the final essay to its origin: typed by the student, pasted from the chatbot, or retyped from it. From this provenance, we derive labels at text and sentence level that distinguish human-written (\textsc{human}), AI-generated (\textsc{ai}) and \textsc{mixed} text, as well as text written after chatbot use that contains no traceable AI material (\textsc{potential\_influence}, \textsc{pi}). Unlike labels derived from the experimental condition, e.g.\ if students had access to a chatbot or not, these reflect what actually ended up in the essay.

On this data, we evaluate two complementary detectors adapted from top-ranked shared-task systems: a feature-based classifier over interpretable linguistic features \citep{indurthi-varma-2025-tesla} and a classifier over frozen Llama-3.1-8B embeddings \citep{jiao-etal-2025-cmi}. We ask:

\begin{description}
\item[RQ1] How well do state-of-the-art detectors identify AI-authored text in ecologically valid student writing, at text and sentence level? How do they treat text that combines AI output with the student's own writing, and text written after chatbot use that contains no copied or retyped AI output?

\item[RQ2] What do the detectors rely on, and is it robust to changes a student could plausibly make, such as fixing or adding spelling errors, or asking the chatbot to make its text sound like a student's?
\end{description}

\noindent Our contributions are:
\begin{itemize}
\item A benchmark of authentic student essays with provenance-based AI-authorship labels at text and sentence level, including partial and untraceable AI use. We will release the dataset under CC BY 4.0 and code upon publication.\footnote{\url{https://github.com/darius-ipn/pisa-ai-authorship}}
\item An evaluation of two complementary top shared-task detectors on this authentic data.
\item An analysis of the detectors’ robustness to realistic manipulations.
\end{itemize}

\section{Related Work}
Research on AI-generated text in education has asked two questions: how well humans, in particular teachers, can tell AI-generated from student writing, and how well automated detectors can.

\paragraph{Human detection.}
Teachers distinguish AI-generated from student writing only moderately well. Pre-service and experienced teachers recognised fewer than half of GPT-4-generated essays (recall = 45.1\% and 37.8\%), and their confidence was unrelated to their accuracy \citep{fleckenstein2024teachers}. College instructors picked ChatGPT essays from a pair in 70\% of cases (chance: 50\%; \citealp{waltzer2024can}). Both instructors and novice EFL teachers report relying mainly on the absence of errors \citep{waltzer2024can, de2024can}, so that well-written learner texts are sometimes taken for AI. Experiment~3 tests whether automated detectors share this reliance.
Humans thus detect AI-generated text only moderately above chance, if at all, and their general confidence says little about their accuracy, which motivates automated detection. 

\paragraph{Automated detection.}
On Graduate Record Examination essays, \citet{jiang2024detecting} detected \texttt{gpt-3.5-turbo} essays with accuracy above 0.98 and suggested keystroke logs for the harder case of partial AI use, which our dataset captures. The 2025 \emph{AI vs.\ Human} shared task \citep{chowdhury-etal-2025-genai} contrasted IELTS and TOEFL essays with essays generated by seven LLMs, among them \texttt{gpt-4o-mini}, prompted to write like non-native students; the top systems reached macro-F1 above 0.98. We adopt two complementary systems from it. \citet{jiao-etal-2025-cmi} (rank~1 for English, macro-F1 = 0.999) classify mean-pooled hidden states of a frozen Llama-3.1-8B with a small convolutional network; their ablations attribute most of the signal to the LLM representation. \citet{indurthi-varma-2025-tesla} (rank~6, macro-F1 = 0.986) train classical classifiers on interpretable style, complexity, bias, affect and moral features, of which complexity features were the most discriminative.

\paragraph{Robustness.}
Such results could suggest that the task is solved, but detectors are vulnerable to changes in surface form. 
None of 14 tools tested by \citet{weber2023testing} exceeded an accuracy of 80\%, and accuracy dropped to 42\% for manually edited and 26\% for paraphrased ChatGPT output. On student essays, simple perturbations, like synonym replacement, lowered the accuracy of fine-tuned RoBERTa detectors to 0.50--0.60 \citep{peng2023hidding}. Experiments~3 and~4 test related manipulations that a student could plausibly apply.

\paragraph{Our setting.}
In all these studies, the AI texts were prompted by researchers, and texts were either fully human-written or fully AI-generated; \citet{weber2023testing} explicitly left hybrid writing open. In our data, students used the chatbot voluntarily and combined its output with their own writing. Keystroke logs make every character attributable to its origin and allow labels for whole texts and single sentences, including partial (\textsc{mixed}) and untraceable (\textsc{pi}) AI use.

\section{Data}
\label{sec:data}

We use data from a writing study conducted with 2{,}429 15-year-old students in German secondary schools alongside the PISA 2025 assessment, as an addition to the PISA 2025 Foreign Language Assessment \citep{oecd2021pisafla}. Following \citet{schaller-etal-2026-keyscore}, we refer to it as PISA 2025 FLA. The data comprise 4{,}091 writing sessions on two argumentative tasks (\emph{advertisement} and \emph{teacher}), taken from the MEWS study \citep{KELLER2020100700}.

Each student wrote one essay per task --- one with (\emph{Chat}) and one without (\emph{NoChat}) unrestricted access to a \texttt{gpt-4o-mini}-based chatbot for writing support \citep{openai2024gpt4omini}. The assignment of conditions to tasks and the order in which the tasks were completed were randomized. Not all students completed both tasks.
Writing sessions were limited to 30 minutes and logged as an event stream of keystrokes, cursor positions, clipboard operations, and the full text of all chat requests and responses.  

In a naive approach, one might assign all essays from the \emph{Chat} condition and all their contained sentences an \emph{AI} gold standard label and label all \emph{NoChat} material as \emph{human}.
However, as users from the \emph{Chat} group could have chosen not to use the chatbot and not every instance of chatbot use is reflected in the final essay, we assign the final labels in a more informed way by first determining every character's origin and then aggregating this information into sentence and text labels. 1{,}889 of the 1{,}981 \emph{Chat} essays involved chatbot use, but only 1{,}134 contain AI material in the final essay (Appendix~\ref{app:ai-use}).

\subsection{Determining character provenance}
\label{sec:provenance}
Characters entered by keystroke are labelled as \emph{typed}. Pasted material inherits the origin of the most recent copy or cut event: \emph{chat} for chatbot responses, \emph{task} for the task description, and \emph{typed} for text from the essay itself or the student's own chat prompts.

Since the logger occasionally misses copy events, a paste that does not match the clipboard content is compared to all chatbot responses received so far and labelled \emph{chat} if its similarity is at least 0.85; otherwise it is marked as \emph{external}. We measure similarity with \texttt{difflib.SequenceMatcher} \citep{python_difflib} after lowercasing and normalising whitespace; if one text is fully contained in the other, similarity is set to 1.

 


We record two additional kinds of information necessary for the later gold standard label.
First, we record whether each character was inserted before or after the student’s first chatbot interaction, which separates \textsc{human} from \textsc{potential\_influence} text (Section~\ref{sec:datasets}).


Second, as individual events are occasionally dropped by the logger, we use a rule-based heuristic to reconstruct the missing edit from the surrounding text states for such cases; characters originating from these reconstructions are marked as \emph{inferred}, everything else as \emph{observed}. 

\subsection{Gold Labels}
\label{sec:datasets}

\paragraph{Sentence labels.} We segment the final essays and the chatbot responses with pySBD \citep{sadvilkar-neumann-2020-pysbd}, adding a split rule for the frequent omission of whitespace after sentence-final punctuation in learner texts. Each sentence is then labelled based on the character provenance described in Section~\ref{sec:provenance}.

All sentences written before the first chatbot interaction, and all NoChat sentences, are labelled \textsc{human}, unless at least 10\% of their characters come from unresolved pastes (\textsc{unknown\_paste}) or at least 50\% originate from the task description (\textsc{task}).
Thresholds were set manually. Varying them (AI share 80–95\%, unresolved-paste share 5–20\%, task share 30–70\%, similarity 0.80–0.95) changes the labels of at most 1\% of sentences, though it affects the size of the small \textsc{mixed} class.

Post-AI sentences are labelled \textsc{ai} if at least 90\% of their characters are AI-sourced, \textsc{mixed} if this share is above 0\%, and \textsc{potential\_influence} otherwise (see retyping below).

Manual retyping of chatbot output is not visible at character level, as retyped characters are logged as keystrokes. We therefore compare each post-AI sentence without AI characters to all sentences of the session's
chatbot responses and label it \textsc{ai} if the maximum similarity (SequenceMatcher ratio) is $\geq 0.90$ and \textsc{mixed} if it is $\geq 0.85$. This range captures retyped sentences with small edits. Most \textsc{mixed} sentences instead arise from partial pasting.
The remaining sentences are labelled \textsc{unknown\_paste} if their unresolved paste share reaches 10\%, and \textsc{potential\_influence} otherwise. \textsc{Potential\_influence} thus marks material written after the student started using the chatbot, but without obvious copy-pasting or retyping of chatbot responses.

\paragraph{Text labels.} Essays with at least 50\% inferred characters are labelled \textsc{unknown\_logger}, as a high share across a whole essay indicates unreliable logging of the session. 
Otherwise, the following rules apply in order: An essay is labelled \textsc{ai} if at least half of its sentences are \textsc{ai}; otherwise \textsc{mixed} if it contains at least one \textsc{ai} or \textsc{mixed} sentence; otherwise \textsc{unknown\_paste} if it contains at least one \textsc{unknown\_paste} sentence; otherwise \textsc{potential\_influence} if it contains at least one \textsc{potential\_influence} sentence. All remaining essays, i.e.\ those written entirely before chatbot use or in the NoChat condition, are labelled \textsc{human}. \textsc{task} sentences do not affect the text label.

\subsection{Experimental Subset and Data Splits}
\label{sec:subset}

We release the full labelled dataset; for our experiments we use the following subset:

\paragraph{Exclusions.}
We only use material that is not removed by any of the following exclusion criteria and that carries one of the four labels \textsc{human}, \textsc{ai}, \textsc{mixed} or \textsc{potential\_influence}. At text level, we exclude essays with fewer than 10 or more than 10{,}000 characters, essays consisting only of non-lexical material, essays labelled \textsc{unknown\_logger} or \textsc{unknown\_paste}, and essays not written predominantly in English, identified semi-automatically with \texttt{lingua} \citep{lingua} and word-frequency heuristics based on \texttt{wordfreq} \citep{speer2018wordfreq}. At sentence level, we additionally exclude sentences labelled \textsc{task} or \textsc{unknown\_paste} and sentences with at least 50\% inferred characters. Table~\ref{tab:exclusions} in Appendix~\ref{app:addtab} reports the material remaining after each step.

\paragraph{Data splits.}

We split the texts into 80\% train, 10\% development and 10\% test data, assigning all essays of a student to the same partition and balancing labels and writing tasks across partitions.
Sentences inherit their data split assignment from the text they belong to. As texts in the train split were slightly longer, the train split on the sentence level contains more than 80\% of all sentences.

\paragraph{Final dataset.}
The experimental subset comprises 3,303 essays by 2,204
students and 44,133 sentences. It contains 1,728 essays
from the \emph{Chat} and 1,575 from the \emph{NoChat} condition. 1,662 of the essays are based on the
\emph{advertisement} and 1,641 on the \emph{teacher} writing task. The
splits comprise 2,634 / 335 / 334 texts and 36,683 / 3,751 / 3,699 sentences
for train, development, and test.

\begin{table}[t]\centering\small
\setlength{\tabcolsep}{2.5pt} 
\begin{tabular}{lrrrr}
\toprule
& \multicolumn{2}{c}{Sentences} & \multicolumn{2}{c}{Texts}\\
\cmidrule(lr){2-3}\cmidrule(lr){4-5}
Label & complete & subset & complete & subset\\
\midrule
\textsc{human}            & 23{,}560 & 18{,}756 & 1{,}802 & 1{,}664\\
\textsc{ai}               & 21{,}266 & 17{,}358 & 1{,}009 &   952\\
\textsc{pot.-infl.}       &  8{,}133 &  7{,}290 &   403   &   381\\
\textsc{mixed}            &    872   &    729   &   321   &   306\\
\textsc{unknown\_paste}   &  1{,}963 &    ---   &   152   &   ---\\
\textsc{unknown\_logger}  &    ---   &    ---   &   226   &   ---\\
\textsc{task}             &    151   &    ---   &   ---   &   ---\\
\midrule
Total                     & 55{,}945 & 44{,}133 & 3{,}913 & 3{,}303\\
\bottomrule
\end{tabular}
\caption{Label distribution in the complete release and in the experimental
subset.}
\label{tab:datasets}
\end{table}

\section{Experiments and Results}

We frame AI-authorship detection as supervised classification at text and sentence level, using the same classifiers and splits (Section~\ref{sec:subset}) throughout. Experiment~1 establishes baselines and examines the ambiguous classes \textsc{mixed} and \textsc{pi}; Experiments~2--4 examine what the detectors rely on, through feature analysis (Exp.~2) and surface manipulations that could evade detection (Exp.~3 and~4).

\subsection{Models}
We explore two complementary systems that ranked among the top in the 2025 Shared Task: a feature-based classifier over interpretable linguistic features and a classifier over LLM embeddings. Hardware, compute, model versions and licenses, and software versions are given in Appendix~\ref{app:impl}.
\paragraph{Feature-based.}
Following \citet{indurthi-varma-2025-tesla}, we represent each text with the NELA feature set \citep{gruppi2020nelagt2019largemultilabellednews}, comprising the groups Style (50 features), Complexity (7), Bias (8), Affect (9) and Moral (11), for a total of 85 features. On these we train Logistic Regression, Random Forest, Extra Trees and XGBoost \citep{chen2016xgboost} with scikit-learn default hyperparameters \citep{scikit-learn} and a fixed random seed, deliberately reproducing the original setup. 

\paragraph{LLM-embedding-based.}
Following \citet{jiao-etal-2025-cmi}, we use a frozen Llama-3.1-8B \citep{grattafiori2024llama3herdmodels} as a proxy model to obtain a single representation per text: input tokens are truncated to a maximum length of 1024, the last-layer hidden states are averaged across all tokens, and the resulting 4096-dimensional vector is classified by a Refined Feature Classification Network (RFCN), a compact convolutional network with three 1D convolutional layers (32, 64, 96 filters; kernel sizes 24, 16, 8) followed by three fully connected layers. The LLM is not fine-tuned; only the RFCN is trained, with AdamW, a linear warmup-decay schedule, batch size 128, learning rate 3e-4, dropout 0.1, and 20 epochs, selecting the checkpoint with the highest macro-F1 on the development set. For hyperparameters not specified in the original paper (pooling size, fully-connected layer widths) see Appendix~\ref{app:impl}. Due to compute constraints we load the model in 8-bit precision rather than fp16. RFCN results are from single training runs; repeated runs differed by up to 0.02 macro-F1, so we do not interpret differences of this size between models.
All non-retrained RFCN evaluations in Experiments~1, 3 and~4 use the same model.

\paragraph{Evaluation.}
Our main metric is macro-F1. We additionally report accuracy in Experiment~1 and the recall of the \textsc{ai} class in Experiment~4, where only \textsc{ai} texts are manipulated.

\subsection{Experiment 1: Baselines}
\label{sec:exp1}

Experiment~1 establishes how well the two detectors separate \textsc{ai} from \textsc{human} text. This binary setup is the basis for all following experiments. As a secondary analysis, we examine how the detectors handle the ambiguous classes \textsc{mixed} and \textsc{potential\_influence} (\textsc{pi}), the grey zone of real-world cases where AI use is partial or unattributable: Are these cases flagged as \textsc{ai} by a detector trained on \textsc{ai} and \textsc{human} only, and can they be separated as classes of their own when the detector is trained on them?

For the baseline, we train the classifiers on \textsc{ai} and \textsc{human} only. To test recognition, we evaluate the same models on three further test sets in which the grey-zone items are added and relabelled as \textsc{ai} (\textsc{mixed}, \textsc{pi}, or both). We also apply the binary models to the \textsc{pi} and \textsc{mixed} items alone. Since the models only know \textsc{ai} and \textsc{human}, this forces each item into one of the two, and the resulting split shows whether a model perceives these cases as more \textsc{ai}- or \textsc{human}-like. 

To test learnability, we train classifiers with the grey-zone classes as their own targets, giving a three-way (\textsc{ai}/\textsc{human}/\textsc{mixed} or \textsc{ai}/\textsc{human}/\textsc{pi}) or four-way classifier.

\paragraph{AI vs.\ Human Baseline.}
On the core \textsc{ai}-vs-\textsc{human} task, both the feature-based XGBoost model and the RFCN on frozen Llama-3.1-8B embeddings are effective (Table~\ref{tab:aihuman}). At text level they are near-identical: macro-F1 = 0.95 (XGBoost) and 0.96 (RFCN), accuracy = 0.95 and 0.97. At sentence level they diverge: XGBoost drops to macro-F1 = 0.88, consistent with less evidence per sentence, whereas the RFCN holds at macro-F1 = 0.95. 
Both remain below the scores of the original systems on the artificially generated shared-task essays (macro-F1 = 0.986, \citealp{indurthi-varma-2025-tesla}; macro-F1 = 0.999, \citealp{jiao-etal-2025-cmi}), as expected for authentic, student-initiated AI use.
In the following, we report XGBoost, the best feature-based model, and the RFCN.
See Appendix~\ref{app:addtab} Table~\ref{tab:aihumanfull} for results on the other models.
\begin{table}[t]\centering\small
\setlength{\tabcolsep}{5pt}
\begin{tabular}{@{}lcccc@{}}
\toprule
& \multicolumn{2}{c}{Text} & \multicolumn{2}{c}{Sentence}\\
\cmidrule(lr){2-3}\cmidrule(lr){4-5}
Test set & XGB & RFCN & XGB & RFCN\\
\midrule
\textsc{ai}/\textsc{human} & 0.95 & \textbf{0.96} & 0.88 & \textbf{0.95}\\
\midrule
\multicolumn{5}{@{}l}{\emph{grey-zone items added as \textsc{ai}}}\\
+\,\textsc{mixed}          & 0.90 & 0.89 & 0.86 & 0.94\\
+\,\textsc{pi}             & 0.81 & 0.84 & 0.76 & 0.81\\
+\,\textsc{mixed}, \textsc{pi} & 0.79 & 0.80 & 0.76 & 0.81\\
\bottomrule
\end{tabular}
\caption{Macro-F1 of the binary detectors, trained on \textsc{ai}/\textsc{human} only. Top: \textsc{ai}-vs-\textsc{human} test set; bottom: grey-zone items added to the test set as \textsc{ai}. All models in Table~\ref{tab:aihumanfull}.}
\label{tab:aihuman}
\end{table}

\paragraph{The grey zone: \textsc{mixed} and \textsc{pi}.}
Adding the grey-zone items to the test set as \textsc{ai} lowers performance, more so for \textsc{pi} than for \textsc{mixed} (Table~\ref{tab:aihuman}). XGBoost text-level macro-F1 drops from 0.95 to 0.90 with \textsc{mixed}, to 0.81 with \textsc{pi}, and to 0.79 with both. The RFCN behaves similarly (text-level macro-F1 = 0.80 with both); at the sentence level, it remains slightly ahead of XGBoost (macro-F1 = 0.81 vs.\ 0.76).
Applying the binary models to the grey-zone items (Table~\ref{tab:greyzone}) shows why the detectors rarely flag these items as \textsc{ai}.

Both models classify almost all \textsc{pi} texts as \textsc{human} (94.6\% XGBoost, 94.7\% RFCN), nearly as often as actual \textsc{human} texts (97.0\%, 97.6\%). 
At sentence level, XGBoost classifies more \textsc{pi} as \textsc{ai} (20.6\%), the RFCN hardly more (8.8\%). \textsc{mixed} splits between the two classes (58.1\% \textsc{ai} for XGBoost, 32.3\% for the RFCN at text level; only 31 test texts).

Trained as classes of their own (Table~\ref{tab:greyzone}), the grey-zone classes are hardly separable for XGBoost. Its text macro-F1 falls to 0.49--0.69, while accuracy stays at 0.76--0.87, because the majority classes \textsc{ai} and \textsc{human} carry the accuracy and the added class is rarely predicted (text level: recall = 0.19 for \textsc{mixed}, 0.08 for \textsc{pi}; 89\% of \textsc{pi} texts are still classified as \textsc{human}). Part of this may be due to the small size of the grey-zone classes (306 \textsc{mixed} and 381 \textsc{pi} vs.\ 1{,}664 \textsc{human} texts). 
The RFCN separates them better, leading XGBoost by 0.12 macro-F1 on \textsc{ai}/\textsc{human}/\textsc{mixed} (0.81 vs.\ 0.69) and by 0.13 on the four-way task (0.62 vs.\ 0.49). 
Experiment~1 shows that both detectors separate \textsc{ai} from \textsc{human} text well, but not what this separation rests on.

\begin{table}[t]\centering\small
\setlength{\tabcolsep}{5pt}
\begin{tabular}{@{}lcccc@{}}
\toprule
& \multicolumn{2}{c}{Text} & \multicolumn{2}{c}{Sentence}\\
\cmidrule(lr){2-3}\cmidrule(lr){4-5}
& XGB & RFCN & XGB & RFCN\\
\midrule
\multicolumn{5}{@{}l}{\emph{(a) Binary models: \% predicted as \textsc{ai}, per true class}}\\
\textsc{ai}    & 91.7 & 94.8 & 85.6 & 94.2\\
\textsc{human} &  3.0 &  2.4 & 11.3 &  5.0\\
\textsc{pi}    &  5.4 &  5.3 & 20.6 &  8.8\\
\textsc{mixed} & 58.1 & 32.3 & 52.1 & 45.5\\
\midrule
\multicolumn{5}{@{}l}{\emph{(b) Grey-zone classes as own targets: macro-F1}}\\
\textsc{ai}/\textsc{human}/\textsc{mixed} & 0.69 & \textbf{0.81} & 0.58 & 0.66\\
\textsc{ai}/\textsc{human}/\textsc{pi}    & 0.61 & 0.68 & 0.57 & 0.69\\
all four                                  & 0.49 & 0.62 & 0.43 & 0.52\\
\bottomrule
\end{tabular}
\caption{Grey-zone classes. (a) Predictions of the binary \textsc{ai}-vs-\textsc{human} models; (b) multiclass detectors trained with \textsc{mixed} and/or \textsc{pi} as separate classes.}
\label{tab:greyzone}
\end{table}



\subsection{Experiment 2: Which features carry the signal?}
\label{sec:exp2}

Experiment~2 asks which features the feature-based detector relies on, and whether they capture general differences between AI and student writing or only differences specific to one writing task. We restrict this analysis to XGBoost, as the dimensions of the RFCN's embeddings have no interpretable meaning. We group the NELA features into \emph{form} (Style, Complexity), which describes how a text is written, and \emph{content} (Affect, Bias, Moral), which is lexicon-based and tied to what a text says. Content features are the likelier source of task-specific signal: On a given task, the chatbot and the students may write about different aspects of the topic, and hence use different vocabulary, without this difference carrying over to another task.

\paragraph{Feature importance.}
We explain the text-level XGBoost detector from Experiment~1, trained on both tasks, with SHAP values \citep{lundberg2017unified}. For every test essay, SHAP assigns each feature a value indicating how strongly and in which direction it pushed the prediction towards \textsc{ai} or \textsc{human}. We report the mean absolute SHAP value per feature as a share of the total importance, both per feature and summed per group.

Form features account for 74.4\% of the detector's SHAP importance (Style 41.7\%, Complexity 32.7\%), content features for 25.6\% (Affect 12.1\%, Bias 9.9\%, Moral 3.6\%; Table~\ref{tab:exp2}). The importance is concentrated on a few features: average word length alone accounts for 19.5\%, followed by punctuation (6.1\%) and function words (4.7\%); the ten most important features together account for 51.6\%. The detector predicts \textsc{ai} for texts with longer words, more punctuation, higher readability indices (SMOG, LIX), more plural nouns and more positive words, and \textsc{human} for texts with more function words, adverbs and verbs. 
In short, \textsc{ai} text is lexically more complex, while student text is shorter and simpler.

\paragraph{Cross-task transfer.}
To test whether these features generalise across tasks, we train XGBoost on one writing task and evaluate it on the same task (\emph{within}) and on the other task (\emph{cross}). If a feature captures a task-independent difference between \textsc{ai} and \textsc{human} writing, what is learned on one task should carry over to the other; if the difference arises from the task, for example because chatbot and students address different aspects of it, performance drops. Both tasks contain a similar share of \textsc{ai} texts (35\% and 38\% in the training set), and training sets are downsampled to equal size (1{,}030 texts) with label proportions preserved. We train on each NELA group, on form and content combined, and on the full NELA set, and report macro-F1 within and cross, each averaged over both tasks, as well as the drop from within to cross for each transfer direction.

Within a task, all feature sets perform similarly (macro-F1 = 0.91--0.95), but they transfer differently (Table~\ref{tab:exp2}). Content features drop the most (by 0.17 macro-F1 on average and by 0.29 when trained on \emph{teacher}), so they partly learn the task rather than the authorship. Form features transfer better (Complexity: 0.03, Style: 0.08, Style and Complexity combined: 0.10), with losses mainly when training on \emph{advertisement} and testing on \emph{teacher}, while content fails mainly in the opposite direction. Only the full NELA set is stable in both directions (drop $\leq$ 0.02), as the groups compensate for each other's losses. With about 130 test texts per task, single per-direction differences should be read with caution. Overall, the difference between \textsc{ai} and \textsc{human} writing lies more in form than in content and largely holds across the two tasks.

\begin{table}[t]\centering\small
\setlength{\tabcolsep}{3.5pt}
\begin{tabular}{@{}lrcccc@{}}
\toprule
& & & & \multicolumn{2}{c}{Drop}\\
\cmidrule(lr){5-6}
Feature set & SHAP\,\% & Within & Cross & a$\to$t & t$\to$a\\
\midrule
\multicolumn{6}{@{}l}{\emph{Form}}\\
Style (50)              & 41.7 & 0.92 & 0.85 & 0.10 & 0.05\\
Complexity (7)          & 32.7 & 0.91 & 0.88 & 0.04 & 0.02\\
Style + Complex. (57)   & 74.4 & 0.95 & 0.85 & 0.20 & 0.01\\
\midrule
Content (28)            & 25.6 & 0.91 & 0.75 & 0.05 & 0.29\\
\midrule
Full NELA (85)          & 100  & 0.93 & 0.92 & 0.02 & 0.01\\
\bottomrule
\end{tabular}
\caption{XGBoost: share of SHAP importance and cross-task transfer (macro-F1). Drop: within minus cross per direction (a = \emph{advertisement}, t = \emph{teacher}).}
\label{tab:exp2}
\end{table}

\subsection{Experiment 3: Spelling Manipulation}
\label{sec:exp3}

Experiment~3 tests whether the detectors use spelling errors, the most visible surface difference between student and chatbot text, as a shortcut. No NELA feature measures spelling directly, so misspellings can affect XGBoost only indirectly (e.g.\ through word length or part-of-speech tags), whereas the RFCN processes the raw tokens.

\paragraph{Manipulation.}
We identify misspellings automatically following \citet{horbach-etal-2017-influence}, accepting words found in a dictionary, frequent in English, or manually verified as topic words or names, and correct them with SymSpell (details in Appendix~\ref{app:spelling}). Student texts contain far more misspellings than chatbot output (4.36 vs.\ 0.72 per 100 words in the text-level test set). In \textsc{correct}, we correct all misspellings in \textsc{human} texts (4.36 $\to$ 0.01). In \textsc{mistake}, we insert misspellings attested in \textsc{human} texts into \textsc{ai} texts, sampled by frequency, until the error rate matches human writing (0.72 $\to$ 5.06). We evaluate the baseline detectors on each manipulated test set and additionally retrain them on each manipulated variant.

\paragraph{Results.}
As expected, neither manipulation affects XGBoost: macro-F1 stays within 0.01 of the baseline throughout (Table~\ref{tab:exp3}). The RFCN loses 0.02 macro-F1 at text level and up to 0.03 at sentence level (\textsc{mistake}: 0.95 $\to$ 0.92). These drops are small but significant in a paired bootstrap over students (Appendix~\ref{app:spelling}). They are also one-directional: inserted misspellings turn \textsc{ai} predictions into \textsc{human} (86 sentences, 0 reverse), and corrections turn \textsc{human} into \textsc{ai} (19 sentences, 1 reverse). Retrained on manipulated data, the RFCN performs as well as before. Spelling, a cue teachers rely on, is thus no shortcut for XGBoost and only a weak cue for the RFCN.

\begin{table}[t]\centering\small
\setlength{\tabcolsep}{4pt}
\begin{tabular}{@{}lcccc@{}}
\toprule
& \multicolumn{2}{c}{Text} & \multicolumn{2}{c}{Sentence}\\
\cmidrule(lr){2-3}\cmidrule(lr){4-5}
Train / test & XGB & RFCN & XGB & RFCN\\
\midrule
orig / orig                         & 0.95 & 0.96 & 0.88 & 0.95\\
orig / \textsc{correct}             & 0.95 & 0.95 & 0.88 & 0.94\\
orig / \textsc{mistake}             & 0.94 & 0.94 & 0.88 & 0.92\\
\textsc{correct} / \textsc{correct} & 0.94 & 0.95 & 0.88 & 0.94\\
\textsc{mistake} / \textsc{mistake} & 0.95 & 0.96 & 0.88 & 0.94\\
\bottomrule
\end{tabular}
\caption{Macro-F1 under the spelling manipulation. Train / test data: original (orig) or manipulated (\textsc{correct}, \textsc{mistake}).}
\label{tab:exp3}
\end{table}

\subsection{Experiment 4: Rewriting AI text to sound like a student}
\label{sec:exp4}

Experiment~4 asks whether students could have hidden their AI use by asking the chatbot to make its text sound like their own writing. Whereas Experiment~3 manipulated spelling, we target the form features identified in Experiment~2. We rewrite all \textsc{ai} texts of the text-level test sets with \texttt{gpt-4o-mini}, the model available to the students, using two prompts. The \textsc{student} prompt is a simple request of the kind a student might write. The \textsc{informed} prompt additionally lists the properties that distinguish \textsc{human} from \textsc{ai} text in Experiment~2. As it uses knowledge learned by the detector, it serves as a best case for evasion rather than a realistic scenario. To test whether the results depend on the rewriting model, we repeat the rewriting with Qwen3-14B \citep{yang2025qwen3} (4-bit, via Ollama\footnote{\url{https://ollama.com}}), which belongs to a different model family than both \texttt{gpt-4o-mini} and the Llama embeddings of the RFCN. All generations use temperature~0; \textsc{human} texts are not changed. Both prompts are given in Appendix~\ref{app:prompts}.

\paragraph{Manipulation check.}
Both \texttt{gpt-4o-mini} prompts moved the form features towards human writing: median average word length fell from 5.69 to 4.45 (\textsc{student}) and 4.39 (\textsc{informed}; human: 4.35), and both rewrites use more pronouns and function words. Only the \textsc{informed} prompt shortened the texts (median word count: 325 $\to$ 183; human: 163), and neither reduced punctuation nor positive sentiment. The rewrites remain semantically close to the originals (mean cosine similarity based on all-MiniLM-L6-v2 text embeddings \citep{reimers-2019-sentence} = 0.84 for \textsc{student}, 0.79 for \textsc{informed}).

\paragraph{Evaluation.}
We evaluate the Experiment~1 models on the rewritten test sets without retraining and report macro-F1 and \textsc{ai} recall, i.e.\ the share of rewritten \textsc{ai} items still detected. 
At sentence level, rewritten texts are re-segmented with pySBD; since they cannot be paired with the original sentences, all are labelled \textsc{ai} (including any the student had written), and we report \textsc{ai} recall on them.

\paragraph{Results.}
Rewriting substantially lowers XGBoost detection (Table~\ref{tab:exp4}): with \texttt{gpt-4o-mini}, its \textsc{ai} recall falls from 0.92 to 0.60 (\textsc{student}) and 0.39 (\textsc{informed}). Even the \textsc{student} prompt, written without knowledge of the detector, hides about a third of detected \textsc{ai} texts. Qwen3-14B rewrites lower recall as much or more. The detector rests largely on form differences that a single prompt can remove.

The RFCN, in contrast, is unaffected at text level: its \textsc{ai} recall stays at 0.94 or above for both prompts and both rewriting models (original: 0.95). 
This is unlikely to be merely recognition of \texttt{gpt-4o-mini}'s style, which produced all \textsc{ai} training data: texts rewritten by Qwen3, from a different model family, are detected just as reliably. The RFCN thus captures properties of LLM-generated text that survive style-level rewriting; we leave it to future work to determine which properties these are.

\paragraph{Sentence level.}
At sentence level, rewriting affects both detectors more strongly. The \textsc{ai} recall of XGBoost falls from 0.86 to 0.25--0.49 across rewrites. Unlike at text level, the RFCN is also clearly affected, though less than XGBoost: its \textsc{ai} recall falls from 0.94 to 0.59--0.90.  
Here, the rewriting model matters about as much as the prompt: Qwen3 lowers \textsc{ai} recall more than \texttt{gpt-4o-mini} for both prompts (\textsc{student}: 0.74 vs.\ 0.90; \textsc{informed}: 0.59 vs.\ 0.70).
As \textsc{ai} recall on rewritten sentences varied by up to 0.05 between training runs, only larger differences should be interpreted. The stronger effect of Qwen3 may indicate that part of this signal is specific to \texttt{gpt-4o-mini}, or that Qwen3 rewrites more strongly.

\begin{table}[t]\centering\small
\setlength{\tabcolsep}{5pt}
\begin{tabular}{@{}lcccc@{}}
\toprule
& \multicolumn{2}{c}{XGBoost} & \multicolumn{2}{c}{RFCN}\\
\cmidrule(lr){2-3}\cmidrule(lr){4-5}
Test set & Rec.\ \textsc{ai} & F1 & Rec.\ \textsc{ai} & F1\\
\midrule
\multicolumn{5}{@{}l}{\emph{Text level}}\\
original                   & 0.92 & 0.95 & 0.95 & 0.96\\
4o-mini \textsc{student}   & 0.60 & 0.81 & 0.99 & 0.98\\
4o-mini \textsc{informed}  & 0.39 & 0.69 & 0.97 & 0.98\\
Qwen3 \textsc{student}     & 0.41 & 0.70 & 0.94 & 0.96\\
Qwen3 \textsc{informed}    & 0.33 & 0.65 & 0.97 & 0.97\\
\midrule
\multicolumn{5}{@{}l}{\emph{Sentence level}}\\
original                   & 0.86 & 0.88 & 0.94 & 0.95\\
4o-mini \textsc{student}   & 0.49 & 0.69 & 0.90 & 0.92\\
4o-mini \textsc{informed}  & 0.29 & 0.57 & 0.70 & 0.82\\
Qwen3 \textsc{student}     & 0.35 & 0.61 & 0.74 & 0.84\\
Qwen3 \textsc{informed}    & 0.25 & 0.54 & 0.59 & 0.77\\
\bottomrule
\end{tabular}
\caption{Detection on rewritten \textsc{ai} test items. \\ Rec.\ \textsc{ai}: recall of the \textsc{ai} class; F1: macro-F1.}
\label{tab:exp4}
\end{table}

\section{Conclusion}

We evaluated two state-of-the-art AI-text detectors on authentic student writing, with labels derived from keystroke, clipboard and chat logs.
Regarding RQ1, AI text from loosely prompted, student-initiated chatbot use is not substantially harder to detect than researcher-generated text. Both detectors separate \textsc{ai} from \textsc{human} reliably at text level. 
At sentence level, which is needed to locate AI material in partially AI-written essays, only the embedding-based detector keeps this performance; the feature-based detector drops noticeably. Partial AI use, however, does not fit the binary framing. \textsc{mixed} text falls between the two classes, and the detectors disagree on where it belongs. \textsc{pi} text is read as human-written and stays hard to separate even when trained as a class of its own. 
Regarding RQ2, what the detectors rely on matters more than how well they score. The feature-based detector has mainly learned that AI text is lexically more complex. This difference holds across tasks, but a single rewriting prompt removes it. The embedding-based detector still recognises authorship of rewritten whole texts, regardless of the rewriting model, but loses a substantial share of rewritten single sentences, especially when another model does the rewriting. Spelling errors, a cue teachers report relying on, play almost no role for either detector.
Text-based detection can thus reliably tell whether a whole essay was produced with a chatbot, but locating AI material at sentence level is fragile, as a single rewriting prompt is enough to hide much of it. AI support that leaves no copied or retyped text in the essay, such as using the chatbot for ideas or language help, cannot be identified from the text at all. Future work should identify which properties the embedding-based detector relies on, improve its robustness at sentence level, and examine what kinds of AI use underlie \textsc{pi} text. 


\section*{Limitations}
Our gold labels rely on manually set thresholds (e.g.\ the 90\% AI share and the 0.85/0.90 similarity thresholds for retyped text), and provenance cannot capture AI material that students paraphrased heavily or used only for ideas. All \textsc{ai} text was produced by \texttt{gpt-4o-mini}, and all writers are 15-year-old EFL students from Germany on two argumentative tasks, so results may not transfer to other generators, populations or genres. The text-level test set is small (96 \textsc{ai}, 169 \textsc{human} texts). Finally, our manipulations are simulated; real students may evade detection in other ways, such as editing the output by hand.

\section*{Ethical considerations}
AI-authorship detectors should not be used as standalone evidence of academic misconduct or for consequential decisions about individual students. False positives may lead to unjustified accusations, and our results show substantial uncertainty for mixed, rewritten, and post-AI writing.
The PISA-FLA data were accessed under the data-protection and research-use procedures of the original study; analyses use pseudonymized data and do not attempt to identify individual students.
Data collection followed the standard PISA consent procedures and was approved by an ethics committee and by the ministries of education of all
participating federal states.
Artificial Intelligence has been used for creating this work in tasks as supporting code writing, translation and editing of text. All generated code was tested before usage.


\bibliography{custom}

\clearpage
\appendix

\section{Additional Tables}
\label{app:addtab}

\begin{table}[h!]\centering\small
\setlength{\tabcolsep}{4pt}
\begin{tabular}{@{}lrr@{}}
\toprule
Step & Texts & Sentences\\
\midrule
Logged sessions                       & 4{,}091 & ---\\
\multicolumn{3}{@{}l}{\textit{Text level}}\\
$-$ empty / length outliers           & 3{,}913 & 55{,}945\\
$-$ \textsc{unk.} texts$^a$           & 3{,}535 & 50{,}138\\
$-$ non-English                       & 3{,}303 & 46{,}076\\
\multicolumn{3}{@{}l}{\textit{Sentence level}}\\
$-$ $\geq 50\%$ inferred              & 3{,}303 & 44{,}358\\
$-$ \textsc{task} / \textsc{unk.\_paste} & 3{,}303 & 44{,}133\\
\bottomrule
\end{tabular}
\caption{From logged sessions to the experimental subset; each row reports the material remaining after the step. $^a$Texts labelled \textsc{unknown\_logger} or \textsc{unknown\_paste}.}
\label{tab:exclusions}
\end{table}

\begin{table}[h!]\centering\small
\setlength{\tabcolsep}{5pt}
\begin{tabular}{@{}lcccc@{}}
\toprule
& \multicolumn{2}{c}{Text} & \multicolumn{2}{c}{Sentence}\\
\cmidrule(lr){2-3}\cmidrule(lr){4-5}
Model & F1 & Acc & F1 & Acc\\
\midrule
Logistic Regression & 0.94 & 0.95 & 0.85 & 0.85\\
Random Forest       & 0.93 & 0.94 & 0.87 & 0.87\\
Extra Trees         & 0.93 & 0.94 & 0.87 & 0.87\\
XGBoost             & 0.95 & 0.95 & 0.88 & 0.88\\
RFCN                & \textbf{0.96} & \textbf{0.97} & \textbf{0.95} & \textbf{0.95}\\
\bottomrule
\end{tabular}
\caption{Core \textsc{ai}-vs-\textsc{human} detection, macro-F1 and accuracy.
All five models are shown; the grey-zone and multiclass analyses below use the
two headline models (XGBoost, RFCN).}
\label{tab:aihumanfull}
\end{table}

\section{Rewriting Prompts}
\label{app:prompts}

Table~\ref{tab:prompts} shows the two prompts used in Experiment~4 (Section~\ref{sec:exp4}). Both were used unchanged with \texttt{gpt-4o-mini} and Qwen3-14B.

\begin{table}[ht]\centering\footnotesize
\setlength{\tabcolsep}{3pt}
\begin{tabularx}{\columnwidth}{@{}l>{\raggedright\arraybackslash}X@{}}
\toprule
\textsc{student} & \emph{Rewrite the following text so that it sounds like it was written by a 15-year-old German student in English class. Only output the rewritten text.}\\
\midrule
\textsc{informed} & \emph{Rewrite the following essay so that it reads like the work of a 15-year-old German student writing in English as a foreign language. Keep the same position and main arguments, but change the style:} \textbullet~\emph{Make it considerably shorter, about 150--200 words.} \textbullet~\emph{Use short, common words instead of long or formal ones.} \textbullet~\emph{Use simple sentences and everyday vocabulary; repeat words rather than using synonyms.} \textbullet~\emph{Write in a personal way, with many pronouns (I, we, you, they) and verbs rather than nouns.} \textbullet~\emph{Use little punctuation: mostly full stops and a few commas, no semicolons, colons, dashes or quotation marks.} \textbullet~\emph{Keep a neutral tone; avoid enthusiastic or strongly positive wording.} \textbullet~\emph{Do not add spelling or grammar mistakes.} \emph{Only output the rewritten essay.}\\
\bottomrule
\end{tabularx}
\caption{Rewriting prompts; the essay is appended to each prompt.}
\label{tab:prompts}
\end{table}

\section{Spelling Manipulation (Experiment 3)}
\label{app:spelling}
A word counts as correctly spelled if it is listed in the Hunspell \texttt{en\_US} dictionary\footnote{\url{https://hunspell.github.io}} or is frequent in English according to \texttt{wordfreq} \citep{speer2018wordfreq} (Zipf frequency $\geq 3.5$, i.e.\ about three occurrences per million words); the latter accepts common British spellings, which \texttt{en\_US} rejects.
Topic words and names are often missing from both resources and would be flagged as errors. We therefore also accept words from our corpus, under conditions that keep out recurring learner misspellings: a word must occur in at least five texts and under at least two labels, so that a single student repeating a misspelling does not count as evidence; it must not be German; and it must not be one edit away from a frequent English word (Zipf $\geq 4$), unless it is mostly capitalised, i.e.\ a name. This yields 290 candidate types, of which we manually accepted 32 as correct spellings (mainly topic words and names).
Frequent German words and contractions without an apostrophe (\emph{dont}) are not counted as errors. Frequent misspellings were corrected by hand at type level, all others with the nearest dictionary word suggested by SymSpell\footnote{\url{https://github.com/wolfgarbe/SymSpell}} (Damerau--Levenshtein distance $\leq 3$). In the text-level test set, \textsc{correct} alters 151 of 169 \textsc{human} texts and \textsc{mistake} alters 95 of 96 \textsc{ai} texts; the \textsc{mistake} error rate overshoots human writing because of errors already present in the \textsc{ai} texts.

Table~\ref{tab:exp3-sig} compares the RFCN under each spelling scenario of Experiment~3 (Section~\ref{sec:exp3}) with the unmanipulated baseline on the same test items.

\begin{table}[ht]\centering\small
\setlength{\tabcolsep}{3pt}
\begin{tabular}{@{}lccc@{}}
\toprule
Train / test & $\Delta$F1 & 95\% CI & Flips\\
\midrule
\multicolumn{4}{@{}l}{\emph{Text level}}\\
orig / \textsc{correct}             & $-$0.016 & [$-$0.032, $-$0.004]* & 0 / 4\\
orig / \textsc{mistake}             & $-$0.025 & [$-$0.048, $-$0.008]* & 6 / 0\\
\textsc{correct} / \textsc{correct} & $-$0.012 & [$-$0.034, \phantom{$-$}0.008] & 2 / 5\\
\textsc{mistake} / \textsc{mistake} & $-$0.004 & [$-$0.021, \phantom{$-$}0.013] & 3 / 2\\
\midrule
\multicolumn{4}{@{}l}{\emph{Sentence level}}\\
orig / \textsc{correct}             & $-$0.006 & [$-$0.012, $-$0.002]* & 1 / 19\\
orig / \textsc{mistake}             & $-$0.028 & [$-$0.037, $-$0.020]* & 86 / 0\\
\textsc{correct} / \textsc{correct} & $-$0.003 & [$-$0.010, \phantom{$-$}0.003] & 20 / 37\\
\textsc{mistake} / \textsc{mistake} & $-$0.005 & [$-$0.012, \phantom{$-$}0.003] & 70 / 32\\
\bottomrule
\end{tabular}
\caption{RFCN under the spelling manipulation, compared with orig / orig on the same test items (265 texts, 3{,}067 sentences). $\Delta$F1: difference in macro-F1. 95\% CI: paired bootstrap over students (2{,}000 resamples); * = CI excludes zero. Flips: items whose prediction changed, \textsc{ai}$\to$\textsc{human} / \textsc{human}$\to$\textsc{ai}.}
\label{tab:exp3-sig}
\end{table}

\section{Chatbot Use in the Chat Condition}
\label{app:ai-use}

Table~\ref{tab:release-stats} shows how often students in the \emph{Chat} condition used the chatbot and how often its output ended up in the final essay (Section~\ref{sec:data}).

\begin{table}[ht]\centering\small
\setlength{\tabcolsep}{4pt}
\begin{tabular}{@{}lrr@{}}
\toprule
& Texts & Sentences\\
\midrule
\emph{Chat} condition            & 1{,}981 & 31{,}936\\
\quad chatbot used               & 1{,}889 & ---\\
\quad AI material in final essay & 1{,}134 & ---\\
Retyping suspects                & ---     & 572\\
\bottomrule
\end{tabular}
\caption{Chatbot use in the \emph{Chat} condition of the complete release (3{,}913 texts, before exclusions). Retyping suspects: post-AI sentences without AI-sourced characters that match a chatbot sentence with similarity $\geq 0.85$.}
\label{tab:release-stats}
\end{table}

\section{Implementation Details}
\label{app:impl}

\paragraph{Models.} We use the base model Llama-3.1-8B (Llama 3.1 Community License), loaded in 8-bit precision with bitsandbytes due to GPU memory constraints. As no code was released, we reimplemented the RFCN following \citet{jiao-etal-2025-cmi}; for unspecified hyperparameters we use max-pooling (size 2) after each convolutional layer and fully connected layers of width 128 and 64 before the output layer, with ReLU activations and dropout before and after the first fully connected layer. We use a 10\% linear warmup, AdamW with default weight decay (0.01), unweighted cross-entropy, and seed 42. Embeddings are mean-pooled over non-padding tokens and stored in float16.
For Experiment~4 we use 
\texttt{gpt-4o-mini} (\texttt{gpt-4o-mini-2024-07-18}, OpenAI Terms of Use) and Qwen3-14B (14.8B parameters,
Q4\_K\_M quantization, Ollama 0.40.2, thinking disabled, Apache 2.0). No hyperparameter search was performed.

\paragraph{Compute.} All local experiments ran on a single NVIDIA RTX 3060 (12\,GB). Extracting Llama-3.1-8B embeddings, training all RFCN variants and rewriting with Qwen3-14B took approximately 6 GPU hours in total, most of it for embedding extraction. Rewriting with \texttt{gpt-4o-mini} used the OpenAI API.

\paragraph{Software.} Table~\ref{tab:software} lists software versions.

\begin{table}[h]\centering\footnotesize
\setlength{\tabcolsep}{4pt}
\begin{tabular}{@{}lll@{}}
\toprule
Package & Version & Used for\\
\midrule
Python                & 3.11.16 & \\
nela-features         & 3.0.1   & NELA features\\
scikit-learn          & 1.7.1   & LR, RF, ET\\
xgboost               & 3.2.0   & XGBoost\\
shap                  & 0.51.0  & Exp.~2\\
pysbd                 & 0.3.4   & Segmentation\\
lingua                & 2.1.1   & Language ID\\
wordfreq              & 3.1.1   & Language ID, spelling\\
symspellpy            & 6.10.0  & Exp.~3\\
sentence-transformers & 6.1.0   & Exp.~4\\
torch                 & 2.5.1   & RFCN\\
transformers          & 5.17.0  & Llama embeddings\\
accelerate            & 1.15.0  & Model loading\\
bitsandbytes          & 0.50.2  & 8-bit loading\\
Ollama                & 0.40.2  & Qwen3 (Exp.~4)\\
\bottomrule
\end{tabular}
\caption{Software versions.}
\label{tab:software}
\end{table}


\end{document}